\pdfoutput=1

\documentclass[11pt]{article}

\usepackage{emnlp2021}

\usepackage{times}
\usepackage{latexsym}
\usepackage{amsmath}
\usepackage{graphicx}
\usepackage{bm}
\usepackage{tabularx}
\usepackage[font=small,skip=4pt]{caption}
\usepackage{placeins}
\usepackage{multirow}

\usepackage[T1]{fontenc}

\usepackage[utf8]{inputenc}

\usepackage{microtype}
\usepackage{amsmath, amsthm, amssymb, amsfonts}

\newcommand{\Sref}[1]{\S\ref{#1}}
\newcommand{\yellow}[1]{{\setlength{\fboxsep}{0pt}{\colorbox{yellow}{#1}}}}

\title{Improving Span Representation for Domain-adapted Coreference Resolution }

\author{Nupoor Gandhi\textsuperscript{\dag}, Anjalie Field\textsuperscript{\dag}, Yulia Tsvetkov \textsuperscript{\S}\\
  Carnegie Mellon University \textsuperscript{\dag}, University of Washington \textsuperscript{\S} \\
  \texttt{\{nmgandhi,anjalief\}@cs.cmu.edu,yuliats@cs.washington.edu} }

\begin{document}
\maketitle
\begin{abstract}
Recent work has shown fine-tuning neural coreference models can produce strong performance when adapting to different domains. However, at the same time, this can require a large amount of annotated target examples. In this work, we focus on supervised domain adaptation for clinical notes, proposing the use of concept knowledge to more efficiently adapt coreference models to a new domain. We develop methods to improve the span representations via (1) a retrofitting loss to incentivize span representations to satisfy a knowledge-based distance function and (2) a scaffolding loss to guide the recovery of knowledge from the span representation. By integrating these losses, our model is able to improve our baseline precision and F-1 score. In particular, we show that incorporating knowledge with end-to-end coreference models results in better performance on the most challenging, domain-specific spans\footnote{Code publicly available at \url{https://github.com/nupoorgandhi/i2b2-coref-public}}.
\end{abstract}

\section{Introduction}
Recent work has achieved high performance on coreference resolution in standard benchmark datasets like OntoNotes \cite{kirstain-etal-2021-coreference, joshi2020spanbert,AB2/MKJJ2R_2013}. However, in many real world settings where coreference resolution would be valuable, text differs greatly from these standard datasets. For example, coreference resolution over clinical notes can enable tracking a patient's progress and treatment history. However, clinical notes contain acronyms and medical terminology. Annotating new training data for every domain of interest is expensive and time-consuming, and coreference models trained on existing benchmark datasets perform worse on other domains \cite{srivastava2020noisy,xu-choi-2020-revealing, joshi2020spanbert}. In this work, we develop a domain-adaptation model for coreference resolution that requires only a small number of target training examples and target domain knowledge.

Our primary approach involves incorporating domain-knowledge into the \textit{span representations} learned by an end-to-end neural system  \cite{lee-etal-2017-end}. A span representation is a vector representation of a contiguous set of tokens. When determining if a given mention refers to an antecedent, span representations are used by the model to (1.) select a set of candidate mentions and (2.) select an antecedent from the candidates for the given mention. Thus, a high-quality span representation encodes the semantic meaning of the span tokens and their local context. \citet{joshi2020spanbert} introduced SpanBERT, a pre-training method extending BERT, designed to improve performance on span-selection tasks that involves masking contiguous spans rather than tokens.
Span representations are derived by concatenating the pre-trained transformer outputs at the boundary tokens with an attention-weighted vector over the span tokens. These representations are fed into a coreference resolution model, thus integrating SpanBERT into an end-to-end coreference resolution system.

SpanBERT is able to capture coreference structure implicitly in rich span representations. 
The expressiveness of the SpanBERT representation is apparent from extrinsic coreference performance, but also through probing tasks that have shown that span representations can capture headedness, coreference arcs, and other linguistic features of coreference \cite{kahardipraja2020exploring}. The best coreference performance and span representations are obtained by training the end-to-end model with SpanBERT using labeled coreference data.

When adapting a coreference model to a new domain, fine-tuning or continued training can greatly improve performance, but this approach can be computationally expensive and requires a large amount of labelled documents from the target domain \cite{gururangan-etal-2020-dont,xia2021moving}.
Neural models  have also been criticized for largely relying on shallow heuristics in the text, suggesting this data-driven learning method requires many target examples to learn a new target distribution \cite{lu2020conundrums, rosenman-etal-2020-exposing}.

The presence of out-of-vocabulary words in a new domain can create additional challenges. SpanBERT uses wordpiece tokenization, which can lead to misleading meaning representation for spans when a single wordpiece belongs to spans with different meanings \cite{joshi2020spanbert, poerner-etal-2020-inexpensive}. Consider for example, the spans \textit{euthmyia} and \textit{dementia}, both of which are common medical terminology but out-of-vocabulary words for SpanBERT, which tokenizes them to contain a common wordpiece: ``\#\#ia''. As described by \citet{poerner-etal-2020-inexpensive}, this can lead to a coreference model incorrectly predicting the spans coreferent, since the suffix ``\#\#ia'' is commonly associated with diseases. A coreference model could correct this by learning a more meaningful representation for the prefix tokens and downweighting the suffix ``\#\#ia'' token, but this would take many target domain training examples.

Instead, we propose a more efficient method for integrating domain-specific knowledge into SpanBERT-based span representations, which requires only a small number of target training samples and leverages domain-specific concept knowledge. 

We take a set of spans with some similarity in meaning to be a concept, and we use concepts of varying granularity (e.g., \textit{problem}, \textit{headache}).  

First, we introduce a \textit{retrofitting loss} (\Sref{sec:retro}) which guides the representation learning of span pairs to satisfy a knowledge-based distance function. This distance function reflects pairwise span relationships from our concept knowledge. As a result, we are able to align the target domain coreference structure encoded in the span representation with the global meanings of the spans. This allows the end-to-end model to more efficiently build more meaningful span representations for the target domain.

We also introduce an auxiliary \textit{scaffolding loss} (\Sref{sec:scaf}) for a concept prediction task in order to ensure that knowledge relevant to the coreference task can be recovered from the span representation. This usage of an auxiliary task to produce a useful inductive bias was introduced in \citet{swayamdipta-etal-2018-syntactic} to add a syntactic labeling loss for coreference resolution since syntactic constituents are often coreferent. Spans belonging to the same concept within our concept knowledge usually corefer, so we generalize this technique to a broad, knowledge-based lexicon in our domain adaptation setting. While our retrofitting loss integrates relative knowledge into the span representation, we are able to supplement the span representation with global meaning using the scaffolding loss.

To evaluate our models, we take OntoNotes as our source domain and the i2b2/VA corpus of clinical notes as our target domain. We train our model on the source domain and 200 examples from the target domain, we evaluate model performance on the target domain. Our knowledge concepts are from the i2b2/VA dataset and the UMLS Metathesaurus.

First, we describe our methodology and introduce our retrofitting and scaffolding loss functions in \Sref{sec:model}. Then, we describe our experiments to quantify model performance on our target domain in \Sref{sec:exp}, and finally we demonstrate the performance improvement over our baseline and on rare/OOV spans \Sref{sec:results}.

The main contribution of this work is in developing methods to integrate concept knowledge into coreference resolution systems to improve domain adaptation. We outperform our baseline primarily by improving precision, and in doing so, we demonstrate the utility of a set of knowledge concepts for adapting span representations to a new domain. We show that our model's performance does not deteriorate on highly domain-specific spans containing OOD frequently subdivided vocabulary. 

\section{Model}\label{sec:model}

\begin{figure}[t!]
    \centering
    \includegraphics[width=1\columnwidth]{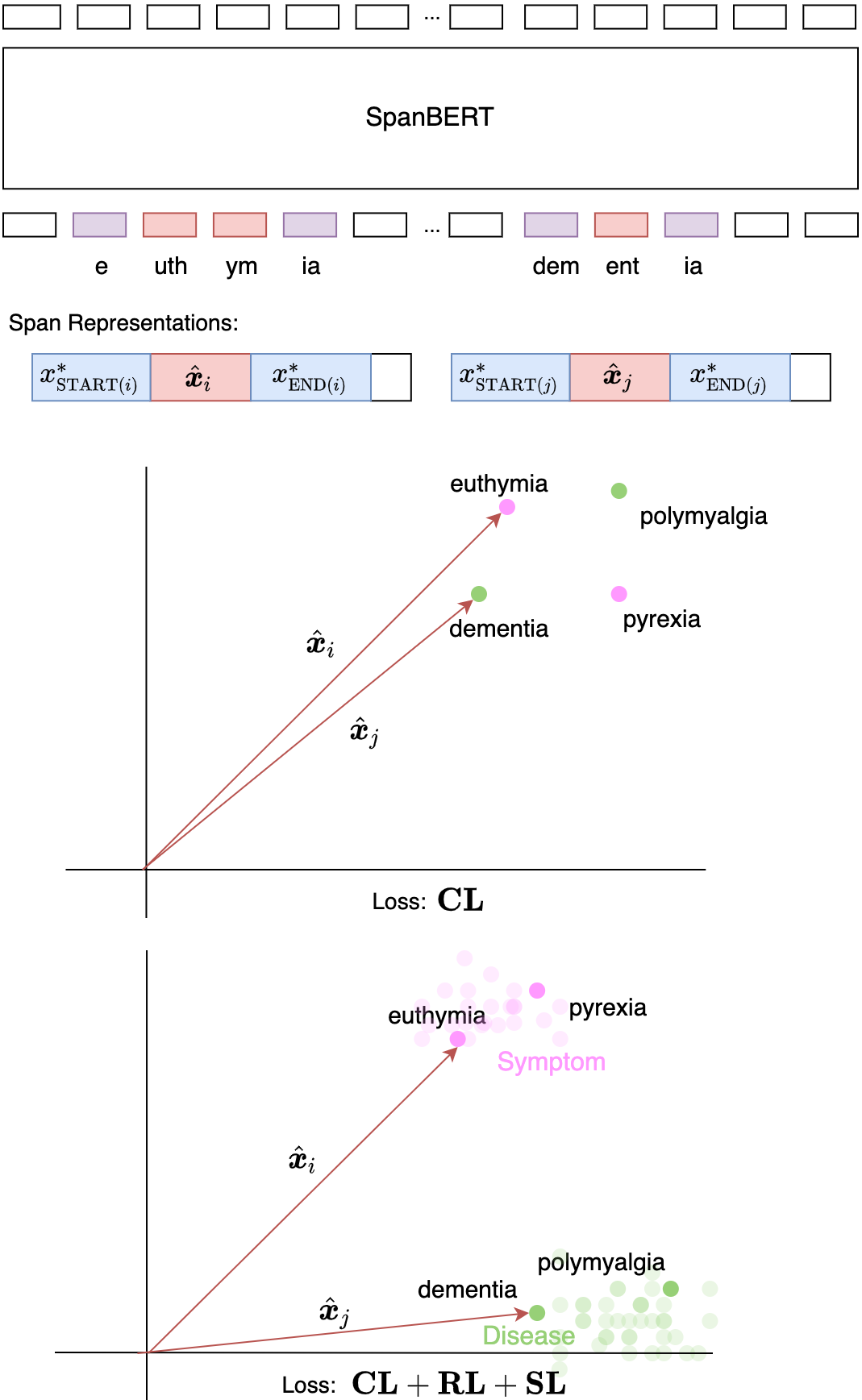}
    \caption{Span representations produced by SpanBERT for mentions ``euthymia'' and ``dementia'' are fed into a co-reference resolution model. The baseline \cite{joshi2020spanbert}, which uses a single coreference loss, $\textbf{CL}$, produces similar span representations for ``euthymia'' and ``dementia'' (top). When we incorporate knowledge concepts, $C_1^k = \{\text{euthymia, pyrexia}, \ldots\}, C^k_2 = \{\text{dementia, polymyalgia}\}, \ldots$, into the span representation using our proposed losses $\textbf{SL}$ and $\textbf{RL}$, the span representations for ``euthmyia'' and ``dementia'' are further apart (bottom), accurately reflecting that ``euthmyia'' is a symptom, while ``dementia''is a disease, even though they share a wordpiece ``ia''.}
    \label{fig:retrofitting_loss}
\end{figure}
The objective of coreference resolution is to identify a set of coreference clusters from a document containing entities, where each coreference cluster contains mentions referencing a single entity. The set of mention candidates are referred to as \textit{spans}, where a span is any contiguous set of tokens in the document.
We can use the coreference clusters $\{C_1, C_2, \ldots, C_m\}$  to define an unconnected graph where the set of spans are the vertices $V = \{s_1, s_2, \ldots, s_n\}$, and there are edges only between spans that belong to the same cluster. 
First, we describe the basic model setup for a neural coreference resolution system, and we then describe our proposed approach, that involves two new losses.

\subsection{End-to-end Coreference Model}
For the set of possible spans, we first produce span representations. The span representation is learned as a part of the neural end-to-end framework introduced in \citet{lee-etal-2017-end}. Given span representations, each span representation $\boldsymbol{h}_i$ is assigned a unary mention score. The mention score reflects the likelihood that a given span is in fact a mention. This score is used to obtain a set of candidate mentions. Each span pair $\boldsymbol{h}_i, \boldsymbol{h}_j$ is assigned a pairwise antecedent score reflecting the likelihood that $\boldsymbol{h}_i$ is the antecedent of  $\boldsymbol{h}_j$. For an arbitrary span pair, the overall score is composed of the antecedent score and the mention scores for each span. The scoring functions are learned using standard feed-forward neural networks, allowing us to derive a distribution over all possible antecedents for a given span $x$.
\begin{align*}
    \textit{P}(y) = \frac{e^{\textit{s}(x,y)}}{\sum_{y'\in\textit{Y}} e^{\textit{s}(x,y')}} 
\end{align*} where $\textit{s}$ is the scoring function as defined in \citet{joshi2020spanbert}. We maximize the likelihood of the correct antecedents from the set of gold mentions, giving us a coreference loss $(\textbf{CL})$:
\begin{align*}
    \textbf{CL} = \log \prod_{i=1}^\textit{N} \sum_{\hat{y} \in \mathcal{Y}(i) \cap \text{GOLD}(i)} \textit{P}(\hat{y})
\end{align*} $\text{GOLD}(i)$ denotes the set of spans in the gold cluster containing span $i$. 
Our baseline span representation is produced using SpanBERT with the single objective of minimizing the coreference loss ($\textbf{CL}$).
\subsection{SpanBERT representation}
SpanBERT is a pre-training method extending BERT that masks contiguous spans and also trains the span boundary representations to predict the
masked span. The span representation $\boldsymbol{h}_i$ is the concatenation of the two SpanBERT transformer states of the span endpoints (first and last word pieces) $\bf{x}_{\text{START}(i)},\bf{x}_{\text{END}(i)}$ and an attention vector $\bf{\hat{x}_i}$ computed over all the word pieces in the span \cite{joshi-etal-2019-bert, joshi2020spanbert}. 
\begin{align*}
    \boldsymbol{h}_i = \left[\bf{x}_{\text{START}(i)},\bf{x}_{\text{END}(i)},\bf{\hat{x}_i},\phi(i)\right]
\end{align*}The attention vector $\bf{\hat{x}_i}$ is intended to best represent the internal span itself (e.g. head word), whereas the endpoints better represent the context \cite{lee-etal-2017-end}. This suggests that the $\bf{\hat{x}_i}$ component of the overall span representation is the most natural part of the span to align with global, non-contextual knowledge.

\subsection{Integrating Knowledge into Span Representation}
We aim to create a span representation such that knowledge can be easily aligned with the coreference structure.
Then, we can learn a span representation geometry such that connected spans are close and disconnected spans are far. In constructing such a vector space, we gain some flexibility to integrate any type of knowledge that shares the same structure as the coreference clusters. We can represent knowledge sources with concept clusters $\{C_1^k, C_2^k, \ldots C_m^k\}$ to resemble coreference clusters, so that we can impose that the coreference cluster graph is consistent with the concept cluster graph via two additional losses.

We propose two complementary approaches to integrating knowledge. A pairwise retrofitting loss is intended to encode relative knowledge and a unary ``scaffold'' loss is intended to encode global knowledge into the span representation.
\subsection{Retrofitting with Concept Knowledge}\label{sec:retro}
We want the span representation $\bf{\hat{x}_i}$ to be close to coreferring spans $\bf{\hat{x}_j}$. Similarly span representations belonging to the same cluster should be close. Similar to \citet{faruqui-etal-2015-retrofitting}, we enforce this objective by defining a custom distance function. But unlike \citet{faruqui-etal-2015-retrofitting}, our custom distance function is task specific and instead of using a lexicon, we are using broad concepts.

\textbf{Concept Knowledge Distance metric}
 We define our distance function to be composed of two elements: coreference information and concept knowledge.
\begin{align*}
    \textit{d}_{\textit{T}}(s_i, s_j) &= \alpha_c d_c(s_i, s_j) 
    + \alpha_k d_k(s_i, s_j)
\end{align*}
Variables $\alpha_c, \alpha_k$ each denote weights that we tune, and $T$ references the document from which valid span pairs are passed into the function. 
 
 \textit{Coreference information} To capture the coreference graph, we recreate the distance between span pairs that do not corefer $(d_c)$. Note that this distance does not discriminate between span pairs that belong to separate coreference clusters and span pairs where one span does not belong to any coreference cluster at all. We define $d_c(s_i, s_j) = 1$ if the spans do not corefer, otherwise $0$.

\textit{Knowledge}
From our knowledge, we can obtain \textit{concepts}, or sets of spans with some level of similarity of non-contextual meaning: $\{C_1^k, C_2^k, \ldots, C_m^k\}$. Coreferent spans refer to the same entity, and as a result, the concept type (e.g. person) must be consistent for any pair of coreferent spans.
Thus, given a set of concepts, we want spans belonging to the same concept to have similar representations. Accordingly, we define $d_k(s_i, s_j) = 0$ if both spans belong to the same concept type, otherwise $1$.

\textbf{Retrofitting Loss (RL)}
We want to create a span representation with a geometry defined by our custom distance function. We can optimize the end-to-end model to satisfy a loss applied to the $\bf{\hat{x}_i}$ component of the span representation:
\begin{align*}
    \textbf{RL} =\sum_{\ell}\frac{1}{|r^{\ell}|}\sum_{i,j}|d_{T^{\ell}}(s_i^\ell, s_j^\ell) - d(\hat{\boldsymbol{x}}_i^{\ell},\hat{\boldsymbol{x}}_j^{\ell})|
\end{align*}
Here $|r^{\ell}|$ denotes the number of span pairs internal to one document, which we use to normalize, $\ell$ identifies the document that span $s_i, s_j$ belongs to, and the function $d$ is cosine distance. 
\subsection{Concept Identification as an Auxiliary Task}\label{sec:scaf}
We introduce a concept identification auxiliary task to guide the model to construct a span representation from which the concept can be recovered.
\citet{swayamdipta-etal-2018-syntactic} introduces the notion of a ``scaffold'' or auxiliary supervised loss function that is related to the primary task. Since coreferring spans nearly always belong to the same concept in our  concept knowledge, concepts are a good choice for a scaffold. By sharing SpanBERT parameters optimizing for the scaffold loss and the overall coreference loss, we are able to encode the concept type in the span representation. 

\textbf{Auxiliary Scaffolding Loss (SL)} Following from \citet{swayamdipta-etal-2018-syntactic}, we assign a distribution over the set of concepts
\[p( s_i \in C \mid \bf{\hat{x}_i}) = \text{softmax}_c \boldsymbol{w}_c\cdot \bf{\hat{x}_i}\] where $\boldsymbol{w}_c$ is a parameter vector associated with a concept $C$. This gives us the loss:
\[\textbf{SL} = \sum_\ell \frac{1}{|r^{\ell}|}\sum_{i}\log p\left(c_i^k\mid \bf{\hat{x}_i}\right) \]

where $c_i^k$ is the concept associated with span $s_i$.

Finally, we optimize a summation of these losses, weighting each loss with a hyperparameter that we tune.
\begin{align*}
    \mathcal{L} = \beta_1 \textbf{CL} + \beta_2 \textbf{RL} + \beta_3 \textbf{SL}
\end{align*}

\section{Experiments}\label{sec:exp}
\subsection{Datasets}
Our target corpus is a medical notes dataset, released as a part of the i2b2/VA Shared-Task and Workshop in 2011 \cite{uzuner20112010}. 
The dataset contains 251 train documents, 51 of which we have randomly selected for development and 173 test documents. Our dev set is used to select some model parameters (e.g., loss function weights $\beta_1, \beta_2, \beta_3$). The average length of these documents is 962.62 tokens with average coreference chain containing 4.48 spans.
For our source domain corpus, we use OntoNotes, with documents on average half as long as the clinical notes and similar average chain length 4.21.

\subsubsection{Knowledge Lexicons}
\noindent\textbf{i2b2 Concepts (i2b2)}:
In addition to coreference chains, the i2b2/VA dataset includes broad concept labels for spans. We focus on four concepts: \textit{person} (e.g. the patient, Dr. X), \textit{treatment} (e.g. abdominal hysterectomy, the procedure), \textit{problem} (e.g.  coronary artery disease, slurred speech), \textit{test} (e.g. MRI, echocardiogram). The i2b2 dataset annotates coreference chains s.t. corefering spans must belong to the same concept. In \autoref{table:i2b2_concepts}, we report how these i2b2 concepts are distributed among the coreference chains.
\begin{table}[!ht]
\centering
\scalebox{0.75}{
\begin{tabular}
{|c | c | c | c | c |} 
\hline
\multirow{2}{*}{i2b2 Concept} & \multicolumn{2}{c|}{Avg. Chain length} & \multicolumn{2}{c|}{\# of Chains} \\\cline{2-5}
& Train  & Test  & Train  & Test 
\\\hline
Problem	&	2.96	&	2.9	&	1704	&	1186	\\\hline				
Test	&	2.31	&	2.51	&	568	&	360	\\\hline				
Person	&	14.16	&	12.54	&	754	&	571	\\\hline				
Treatment	&	2.66	&	2.63	&	1262	&	1063	\\\hline	\end{tabular}}
\caption{Breakdown of i2b2 concepts in coreference chains}
\label{table:i2b2_concepts}
\end{table}

\noindent\textbf{Unified Medical Language System Concepts (UMLS)}: 
The Unified Medical Language System (UMLS) defines concept unique identifier (CUI) codes in the UMLS Metathesuarus tool \cite{bodenreider2004unified}. Each UMLS concept links synonymous spans, so the UMLS concepts are much more fine-grained than those defined in the i2b2 dataset. For example, a CUI for \textit{headache} would map to \{headaches, cranial pain, head pain cephalgia\}. We used string match to assign a UMLS concept to spans in the training set. This resulted in the identification of 3,500 unique CUI's for the spans. We also experimented with using a partial string match to assign UMLS concepts to spans which we refer to as ``UMLS overlap''. 

\subsection{Baseline}
For our supervised domain adaptation approach, we use a familiar approach of training a model on a source domain and tuning this model on a target domain.
We take the current state of the art end-to-end coreference model from \citet{joshi2020spanbert} for our baseline. 

First, we train this SpanBERT-based end-to-end model on OntoNotes using the hyperparameters from \citet{joshi2020spanbert}. Then, we continue training this model
using the target i2b2 training dataset. Continued training has been shown to be effective for coreference resolution in out-of-domain settings \cite{xia2021moving}.

\subsection{Model Setup}
In order to improve the SpanBERT-based span representation, we introduce the i2b2 and UMLS concept knowledge in two ways: we retrofit the span representation to the concept knowledge (\Sref{sec:retro}) and introduce an auxiliary task of concept identification as a scaffold (\Sref{sec:scaf}).

In our implementation, we add the  corresponding retrofitting loss (\textbf{RL}) and scaffolding loss (\textbf{SL}) from these two objectives to the coreference loss (\textbf{CL}) to produce an overall loss which we optimize for. Aside from the difference in the loss function, the training process for our model resembles that of our baseline. We train our model first with the source domain OntoNotes and then continue training on the i2b2 dataset. 

For the retrofitting loss, we experiment with the knowledge lexicons i2b2 and UMLS individually and together. Recall that the knowledge lexicon distance metric relies on two main components: coreference clusters $C_1, C_2, \ldots$ and knowledge clusters $C_1^k, C_2^k, \ldots$. When using i2b2 or UMLS knowledge concepts individually, we experiment with $\alpha_k, \alpha_c$ values between $(0,1]$ at intervals of $.1$, and we found that $\alpha_k = .2$ and $\alpha_c =1.0$ performs best over the dev set. When using i2b2 and UMLS concepts together, we found that assigning more weight ($\alpha_k=.5$) to the broader i2b2 concepts than the UMLS concepts ($\alpha_k = .2$) performs best over the dev set. When training our model on OntoNotes, we do not have the same knowledge lexicon available, so effectively, we have $\alpha_k = 0$ until we begin training on the i2b2 data.

For the concept identification auxiliary task, we use only i2b2 concepts for our knowledge lexicon, since using the fine-grained UMLS concepts would induce 3,500 class labels. Additionally, since the i2b2 knowledge lexicon is not available for OntoNotes, we ignore \textbf{SL} ($\beta_3 = 0$).  

For our model, we choose 
max\_seq\_length of 512, BERT learning rate of
$2e-5$, and task specific  learning rates of $1e-4$. Similar to \citet{joshi2020spanbert}, we fine-tune 20 epochs for OntoNotes and the i2b2 training examples.

\section{Results}\label{sec:results}
For successful domain adaptation, our model aims to demonstrate the value of incorporating concept knowledge. We evaluate overall coreference performance improvement as a result of the span retrofitting and auxiliary concept identification task in \Sref{sec:eval-coref}. Then, we inspect whether representations for highly domain-specific spans are better for the coreference resolution task in our model than in the baseline (\Sref{sec:eval-domspec}).


\begin{table*}[t!]
\centering
\scalebox{0.8}{
\begin{tabular}
{|c| c | c | c | c | c | c | c | c | c | c| c | c | c|} 
\hline\hline
\multirow{2}{*}{Model Losses} & \multirow{2}{*}{Knowledge} & \multicolumn{12}{c||}{Model Performance} 
\\\cline{3-14}
& & \multicolumn{3}{c|}{MUC} & \multicolumn{3}{c|}{B-cubed} & \multicolumn{3}{c|}{CEAFE} & \multicolumn{3}{c|}{averages} 
\\\cline{3-14}
& & R & P & F-1 & R & P & F-1 & R & P & F-1 & R & P & F-1 \\\hline
Baseline (\textbf{CL}) &	NA	& 70.93 & 72.51 & 71.71 & 64.91 & 66.48 & 65.69 & 54.57& 58.44 & 56.44 & 63.47 & 65.81 & 64.61 \\\hline \hline
\textbf{CL} + \textbf{RL} + \textbf{SL}	&	i2b2, UMLS	&	71.15	&	73.64	& \textbf{72.37}	&	65.03	&	67.59	&	\textbf{66.28}	&	54.77	&	60.64	&	\textbf{57.56}	&	63.65	&	68.04	&	\textbf{65.41} \\\hline
\textbf{CL} + \textbf{RL}	&	i2b2, UMLS	&	70.66	&	73.88	&	72.23	&	64.36	&	67.89	&	66.06	&	54.39	&	60.38	&	57.22	&	63.14	&	67.28	&	65.17 \\\hline
\textbf{CL} + \textbf{SL}	&	i2b2	&	70.28 &	74.14 &	72.16 &	64.22 & 68.27	& 66.18 & 54.69 & 60.43 & 57.41	&	63.43	&	67.17	&	65.24\\\hline

\hline	
\end{tabular}}
\captionof{table}{Overall coreference performance for various combinations of loss functions and knowledge resources. Our model surpasses our Baseline ($\textbf{CL}$) largely as a result of an improvement in precision (scores averaged over 6 random seed initializations).}
\label{table:coref-perf}
\end{table*}
\subsection{Coreference Performance}\label{sec:eval-coref}
In \autoref{table:coref-perf}, we see that combining both of the losses  we introduce in this work (\textbf{CL + RL + SL}) improves the model precision by 2.23\% resulting in a .8\% improvement in the F-1 score.

Combining the retrofitting loss and the auxiliary scaffolding loss performs better than using each individually. It is possible that adding the scaffolding loss alone is not as helpful because it does not contain the UMLS knowledge.

While our model does improve precision and the overall F-1 score, recall largely remains constant. The retrofitting loss pushes unrelated spans belonging to different concepts further apart, and consequently we penalize the detection of any valid mention that does not appear in our concept knowledge. Incompleteness of our concept knowledge may be a contributing factor to the lack of recall improvement.

It is expected that precision should be affected most by the additional losses \textbf{SL} and \textbf{RL}, since both 
are designed to integrate knowledge in the pairwise relationships between spans. 

The scaffolding loss guides the model to be able to distinguish spans belonging to different concepts from the span representation, and the retrofitting loss enforces a knowledge-based distance function between spans. Model recall is partly determined by a unary mention score used to identify candidate mentions from spans in our model. This mention score is impacted by our losses, since span representation is taken as input to the scoring function. However, the pairwise knowledge integrated into the span representation is much more useful for selecting an antecedent among a set of candidates for a given span. Consequently, since our loss functions have a bigger impact on the antecedent scoring function, then they will also have a bigger impact on model precision.

\begin{table}[hbt!]
    \centering
    \scalebox{0.7}{
    \begin{tabular}{| c | c | c | c | c | c | c |}\hline
       \multirow{2}{*}{Model Losses} & \multirow{ 2}{*}{Metric} & \multicolumn{4}{c|}{i2b2 Concepts}  \\\cline{3-6}
       & & Person	&	Problem	&	Treatment	&	Test\\\hline
       \multirow{3}{*}{Baseline (\textbf{CL})} & Avg. R & 63.47	&	50.92	&	54.85	&	48.08 \\\cline{2-6}
       & Avg. P & 83.76	&	79.98	&	82.98	&	84.43 \\\cline{2-6}
       & Avg. F-1 & 69.92	&	58.0	&	\textbf{62.24}	&	54.74 \\\hline
       \multirow{3}{*}{ \textbf{CL} + \textbf{RL} + \textbf{SL}} & Avg. R & 62.66	&	51.98	&	53.17	&	48.67 \\\cline{2-6}
       & Avg. P & 86.23	&	83.20	&	86.22	&	86.67 \\\cline{2-6}
       & Avg. F-1 & \textbf{69.98}	&	\textbf{59.56}	&	61.84	&	\textbf{55.62} \\\hline
    \end{tabular}}
    \caption{Performance on coreference chains belonging to a specific concept. Our model outperforms the baseline on more domain-specific spans indicating our model improves domain adaptation \textit{problem} and \textit{test} coreference clusters}
\label{table:concept-perf}
\end{table}

\begin{table*}[hbt!]
\centering
\scalebox{0.8}{
\begin{tabular}
{| c | c || c | c | c | c | c |}\hline
\multirow{2}{*}{Model Losses} & \multirow{2}{*}{Metric} & \multicolumn{5}{c|}{Avg. \# of wordpieces/span range}\\\cline{3-7}
& & [0.0,1.7)	& 	[1.7,3.4)	& 	[3.4,5.1)	& 	[5.1,6.8)	& 	[6.8,8.5) \\\hline \hline
\multirow{3}{*}{Baseline (\textbf{CL})} & Avg. R	& 	47.19	& 	31.46	& 	35.33	& 	27.77	& 	26.49\\\cline{2-7} 
& Avg. P	& 	79.94	& 	64.55	& 	60.63	& 	54.32	& 	49.28 \\\cline{2-7}
& Avg. F-1	& 	54.64	& 	\textbf{38.40}	& 	41.16	& 	35.10	& 	29.81 \\\hline\hline
\multirow{3}{*}{\textbf{CL} + \textbf{RL} + \textbf{SL}} & Avg. R &	47.18	& 	29.41	& 	36.88	& 	30.76	& 	27.93 \\\cline{2-7}
& Avg. P	& 	83.04	& 	67.42	& 	65.31	& 	54.74	& 	60.17 \\\cline{2-7}
& Avg. F-1	& 	55.19	& 	37.39	& 	\textbf{43.57}	& 	\textbf{37.28}	& 	33.78 \\\hline
\multirow{3}{*}{\textbf{CL} + \textbf{RL} + \textbf{SL} (Overlap UMLS)} & Avg. R & 48.68	& 	31.73	& 	38.03	& 	31.73	& 	29.59 \\\cline{2-7}
 & Avg. P	& 	79.22	& 	63.06	& 	57.89	& 	52.26	& 	51.65 \\\cline{2-7}
& Avg. F-1	& 	\textbf{55.50}	& 	38.25	& 	41.86	& 	37.22	& 	33.103 \\\hline
\multirow{3}{*}{\textbf{CL} + \textbf{RL} (Overlap UMLS)} & Avg. R	& 	47.32	& 	29.71	& 	34.67	& 	29.74	& 	36.82 \\\cline{2-7}
 & Avg. P	& 	80.63	& 	64.58	& 	65.58	& 	55.67	& 	56.89 \\\cline{2-7}
& Avg. F-1	& 	54.92	& 	37.03	& 	42.03	& 	37.07	& 	\textbf{38.11} \\\hline

\end{tabular}}
\caption{We take the set of coreference chains s.t. the average number of wordpieces per span falls withing the range, and evaluate model of the subset. We observe that our model outperforms the baseline coreference chains with spans that are subdivided more frequently }
\label{table:subword-perf}
\end{table*}

\begin{table}[hbt!]
\centering
\footnotesize
{
\begin{tabularx}{\columnwidth}{|  X  |}
\hline
\multicolumn{1}{|c|}{TP Coreferent Span Pair Examples}  \\
\hline\hline
 Hereafter, wife noted development of left sided weakness, facial droop, \yellow{slurring of speech} \ldots with past medical history of atrial fibrillation on coumadin, coronary artery disease, hyperlipidemia, dementia with sudden onset left sided weakness,  \yellow{dysarthria}\\
\hline
evaluation and treatment of \yellow{adenocarcinoma} involving the transverse colon and gallbladder \ldots DISCHARGE DIAGNOSIS: 1. \yellow{Metastatic  gallbladder cancer}\\\hline
He is status post a \yellow{hemiarthroplasty} on 10/17/97 \ldots decreased hematocrit prior to his \yellow{humeral fixation surgery}\\\hline
An angiogram was done which disclosed possible \yellow{subsegmental pulmonary emboli of the upper lobes} as well \ldots patient was bolused with intravenous heparin due to concern for \yellow{pulmonary embolism} \\\hline \end{tabularx}}
\caption{Examples of coreferent span pairs missed by Baseline (\textbf{CL}), identified by our model (\textbf{CL} + \textbf{RL} + \textbf{SL}). In these cases, we can see that wordpiece tokenization is likely misleading the baseline model, since the spans in each pair have few wordpieces in common.}
\label{table:tp}
\end{table}

\begin{table}[hbt!]
\centering
\footnotesize
{
\begin{tabularx}{\columnwidth}{|  X  |}
\hline
\multicolumn{1}{|c|}{FP Non-coreferent Span Pair Examples }  \\
\hline\hline
She underwent an \yellow{open laparoscopy} \ldots The patient is now admitted for \yellow{exploratory laparotomy}
\\\hline
Right heart catheterization and \yellow{coronary angiography} on October 15 \ldots urgently transferred by Dr. Lenni Factor for possible \yellow{angioplasty} 
\\\hline
78-yo male with \yellow{atrial fibrillation}\ldots Mechanical mitral valve: \yellow{Anticoagulation} was reversed  
\\\hline
He had a \yellow{cardiac catheterization} performed which  revealed \ldots management after this \yellow{hospitalization} and has done very well 
 \\\hline \end{tabularx}}
\caption{Examples of non-coreferent span pairs correctly missed our model (\textbf{CL} + \textbf{RL} + \textbf{SL}), but identified by Baseline (\textbf{CL}). In these cases, we can see that wordpiece tokenization is likely misleading the baseline model, since the spans in each pair have wordpieces in common. }
\label{table:fp}
\end{table}

\subsection{Performance on Domain-specific Spans}\label{sec:eval-domspec}
In addition to overall performance, we are interested in comparing performance on rare spans that do not occur or occur infrequently in the source domain. 
Specifically, we are interested in addressing performance degradation that can occur as a result of wordpiece tokenization. 

For example, consider a coreference cluster containing spans that are on average subdivided many times. This is an indication that the vocabulary of spans in the cluster is out of domain. The attention weighted vector over the tokens in the spans may be a misleading representation since the token embeddings correspond to shorter, less meaningful subwords. We report in \autoref{table:fp} some examples of span pairs that the baseline model incorrectly predicts as coreferent. 

Consider the spans ``open laparoscopy'' and ``exploratory laporotomy''. Their tokenizations would include the subwords ["lap", "\#\#aro", "\#\#tom", "\#\#y"] and ["lap", "\#\#aro", "\#\#sco", "\#\#py"]. This overlap of the first few subwords might lead the baseline to conclude that the spans are similar in non-contextual meaning and consequently coreferent.  From our fine-grained UMLS knowledge, we know that laparoscopy and laparotomy belong to distinct concepts. Our loss functions $\textbf{RL}, \textbf{SL}$ guide the model to produce disparate span representations since the spans map to distinct concepts. Our model uses knowledge to learn to upweight the wordpieces that are meaningful in context of the target domain (e.g., the suffix tokens in this example). We report examples of challenging span pairs that our model identifies and the baseline fails to identify in \autoref{table:tp}.

To collect quantitative evidence for our model's performance on the most challenging spans, we evaluate how model performance changes as the average number of wordpieces per span increases in \autoref{table:subword-perf}. We can observe that as the average number of wordpieces per span increases beyond $3.4$, we start to see an increasing F-1 performance over the baseline. Similar to the overall results for the entire test set, we see that performance improvements in precision are largely responsible.

Beyond average number of wordpieces per span, we can also use the concept labels annotated in the i2b2 dataset to verify the performance of our model on domain-specific spans. We verified that concepts \textit{person} and \textit{treatment} are respectively dominated by spans like ``doctor'', ``patient'' ``surgery'', and ``procedure'', which are more likely to appear in the source domain than spans like ``afebrile'', ``basal cell carcinoma'', ``asculation'' from the \textit{problem} and \textit{test} concepts. Therefore, the performance improvement in \autoref{table:concept-perf} for concepts \textit{problem} and \textit{test} suggests that our model can outperform the baseline on domain-specific spans. 

\subsection{Visualization}
\begin{figure}[ht!]
    \centering
    \includegraphics[width=0.4\textwidth,height=.8\textwidth]{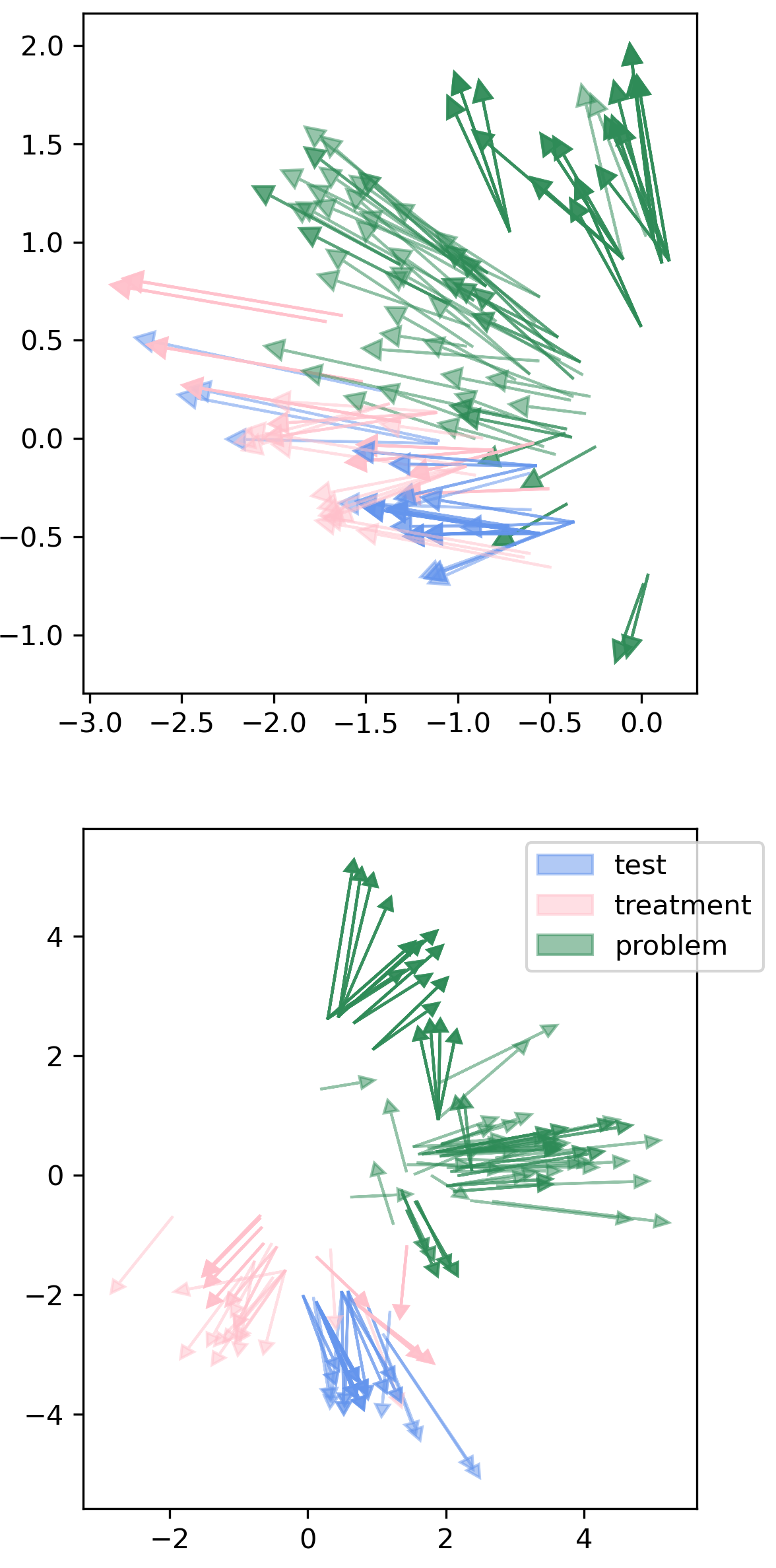}
    \caption{Two-dimensional PCA projections of attention-weighted span tokens for vector pairs holding the mention-antecedent relation for the baseline (top) and our model \textbf{CL + SL + RL} (bottom). Our model uses concept knowledge to construct a span representation that more consistently captures the mention-antecedent relationship specific to each concept.}
    \label{fig:proj}
\end{figure}

We want to visualize how the span representation captures the mention-antecedent relationship for different types of concepts -- specifically how consistently information is extracted from the mention span representation to arrive at a predicted antecedent. We randomly select $200$ span pairs with a mention-antecedent relationship. For each span pair, we extract the attention weighted vector over the span tokens, which is the same piece of the span used to compute \textbf{SL}, \textbf{RL}. For these 768-dimensional vectors, we take the projection vectors from the mention to the antecedent vector. Then, in \autoref{fig:proj} we transform these projections into a 2-dimensional space using PCA and plot them in $\mathbb{R}^2$ similar to \citet{faruqui-etal-2015-retrofitting}. 

For the baseline model (top), most of the mention-antecedent vectors share a similar direction regardless of concept type. However, for the \textbf{CF} + \textbf{RL} + \textbf{SL} model (bottom), there is a clear distinction in mention-antecedent vectors for each concept. This suggests that our model is able to construct a span representation that can capture the mention-antecedent relationship in a way that is (1) specific to the concept (2) consistent across all mention-antecedent pairs belonging to the concept.

\section{Related Work}
Prior to the introduction of end-to-end neural models in \citet{lee-etal-2017-end}, coreference resolution for the clinical domain used pipelined approaches, allowing for the propagation of errors from other core NLU tasks and relying on hand-crafted rules for resolving each type of entity in the domain \cite{jindal2014joint}. Feature-based methods like \cite{jindal2014joint} using knowledge in the coreference model rely on the availability of knowledge at test time. We are focused instead on the case where there may be no concept knowledge available when our model is deployed.  

There has been some limited work in domain adaptation for coreference resolution. 
\citet{yang2012domain} adapts a model trained on the MUC-6 and ACE 2005 datasets to the biomedical domain using an active learning approach, applying data augmentation and pruning techniques.
\citet{zhao2014domain} propose a feature-based active learning method to learn cross-domain knowledge. 
Unlike these works, we take advantage of the modern expressive power of the SpanBERT representation.
With the introduction of SpanBERT, there was a marked performance improvement for several NLU tasks including coreference resolution. \citet{joshi2020spanbert} showed that SpanBERT could be fine-tuned to perform well on several datasets, e.g., GLU and  ACE \cite{wang-etal-2018-glue, doddington2004automatic}.
\\\indent However, \citet{lu2020conundrums} found that SpanBERT coreference resolvers generally rely more on mentions than context, so they are susceptible to small perturbations (e.g., changing all the names/nominal mentions in the test set). More generally, for NLU tasks  several studies have found that neural models rely heavily on shallow heuristics in the text rather than learning the underlying structure of the linguistic phenomenon \cite{peng-etal-2020-learning,rosenman-etal-2020-exposing}, leading to the misinterpretation of context \cite{alt-etal-2020-probing}.

This poses a challenge for adapting a coreference model to noisy domains like clinical text and mostly OOD spans.
With the recent success of fine-tuning for domain adaptation, a natural approach would be to fine-tune the SpanBERT representation, the coreference model, or both \cite{gururangan-etal-2020-dont}.
Fine-tuning the pretrained SpanBERT method alone can be expensive  -- \citet{gururangan-etal-2020-dont} showed that the best performing scheme requires 180K documents. In settings where there are fewer documents available in the target domain, it is still possible to fine-tune a coreference model with SpanBERT \cite{joshi2020spanbert}. 

However, a persisting challenge with adapting the span representation is associated with the wordpiece tokenization employed by SpanBERT. For highly technical domain-specific language, it is natural that there is a higher average number of subwords per span, since it is unlikely that many spans or spans split only once belong to the limited 300,000 word SpanBERT vocabulary. \citet{poerner-etal-2020-e} show that wordpiece tokenization in the biomedical domain can lead to misleading span representations. Purely fine-tuning approaches fail to address this issue, since the SpanBERT vocabulary is constant.

At the same time, SpanBERT implicitly learns a geometry that encodes rich information related to the coreference task.
\citet{hewitt2019structural} show that it is possible to learn a linear projection space from BERT embeddings to capture linguistic phenomena like syntactic dependencies. In \citet{kahardipraja2020exploring}, the authors use a Feedforward Neural Network (FFNN) to probe for properties of coreference structure, finding evidence that the SpanBERT representation can be used to predict coreference arcs and the head word of the span with >90\% F1. Therefore, it is likely that overall performance greatly depends on the quality of span representation.

In our work, we develop multiple techniques to adapt the span representation to a new domain with concept knowledge, allowing the model to be fine-tuned with fewer target domain examples and perform better on highly domain-specific entities. In particular, by incorporating knowledge into the span representation, we are able to restore a global, non-contextual meaning to excessively subtokenized spans.


\section{Conclusion}
We present methods to efficiently adapt coreference resolution models to a new domain using domain-specific concept knowledge. We demonstrate that we can integrate knowledge into the span representation using two losses to (1) retrofit the span representation to the concept knowledge and (2) ensure that knowledge can be recovered from the span representation using an auxiliary concept identification task. Using these methods, we are able to improve the performance of our baseline, especially for highly domain-specific spans. 
\section*{Acknowledgements}
Thanks to Alex Chouldechova, Amanda Coston, David Steier, and the  Allegheny County Department of Human Services for valuable feedback on this work. This work is supported by the Block Center for Technology and Innovation, and A.F. is supported by a Google PhD Fellowship.
\bibliography{anthology,custom}
\bibliographystyle{acl_natbib}

\end{document}